\documentclass{article}

\usepackage[preprint]{neurips_2023}

\usepackage[utf8]{inputenc} 
\usepackage[T1]{fontenc}    
\usepackage{hyperref}       
\usepackage{url}            
\usepackage{booktabs}       
\usepackage{amsfonts}       
\usepackage{nicefrac}       
\usepackage{microtype}      
\usepackage{xcolor}         

\usepackage{natbib}

\usepackage{amsmath}

\usepackage{colortbl}  
\usepackage{graphicx}
\usepackage{graphicx}
\usepackage{subfigure}
\usepackage{multirow}
\usepackage{subfigure}
\usepackage{amsmath}
\usepackage{amsthm}
\usepackage{multirow}

\usepackage{algpseudocode}
\usepackage{algorithm}
\usepackage{physics}
\usepackage{colortbl}
\usepackage{CJKutf8}
\usepackage{caption}
\usepackage{enumitem}
\usepackage{arydshln}
\usepackage{tikz}
\usepackage{pgfplots}

\newcommand{\logofig}{\includegraphics[height=0.98em,trim=0 0.5em 0 0]{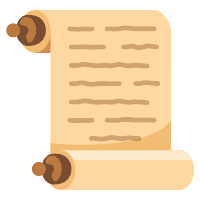}}

\title{LongRecipe\logofig: Recipe for Efficient Long Context Generalization in Large Language Models}

\author{
  Zhiyuan Hu$^{1}$\thanks{Corresponding to: Zhiyuan Hu, \href{mailto:zhiyuan_hu@u.nus.edu}{zhiyuan\_hu@u.nus.edu}. Yuliang Liu contributes equally in this work} \quad Yuliang Liu$^{2}$ \quad Jinman Zhao$^{3}$ \quad Suyuchen Wang$^{4}$ 
  \quad \textbf{Yan Wang}$^{6}$ \quad \\ \textbf{Wei Shen}$^{7}$ \quad 
  \textbf{Qing Gu}$^{2}$ \quad 
  \textbf{Anh Tuan Luu}$^{5}$ \quad
  \textbf{See-Kiong Ng}$^{1}$ \quad
  \textbf{Zhiwei Jiang}$^{2}$ \quad
  \textbf{Bryan Hooi}$^{1}$
  \\
  $^1$ National University of Singapore \quad
  $^2$ Nanjing University \\
  $^3$ University of Toronto \quad
  $^4$ Mila, Qu\'ebec AI Institute / Universit\'e de Montr\'eal \\ \quad
  $^5$ Nanyang Technological University \quad
  $^6$ Tencent Inc \quad
  $^7$ Baidu Inc
}

\begin{document}

\maketitle

\vspace{-6mm}

\begin{abstract}
\vspace{-2mm}
Large language models (LLMs) face significant challenges in handling long-context tasks because of their limited effective context window size during pretraining, which restricts their ability to generalize over extended sequences. Meanwhile, extending the context window in LLMs through post-pretraining is highly resource-intensive.
To address this, we introduce LongRecipe, an efficient training strategy for extending the context window of LLMs, including impactful token analysis, position index transformation, and training optimization strategies. It simulates long-sequence inputs while maintaining training efficiency and significantly improves the model's understanding of long-range dependencies. Experiments on three types of LLMs show that LongRecipe can utilize long sequences while requiring only 30\% of the target context window size, and reduces computational training resource over $85\%$ compared to full sequence training. Furthermore, LongRecipe also preserves the original LLM's capabilities in general tasks. Ultimately, \textit{we can extend effective context window of open-source LLMs from 8k to 128k, achieving performance close to GPT-4 with just one day of dedicated training using a single GPU with 80G memory.}
Our code is released at \href{https://github.com/zhiyuanhubj/LongRecipe}{https://github.com/zhiyuanhubj/LongRecipe}.

\end{abstract}

\vspace{-5mm}
\section{Introduction}
\vspace{-3mm}
LLMs are crucial for NLP and multi-modal tasks. However, they face challenges in applications involving long context, such as in-context learning \citep{brown2020language}, long document summarization \citep{koh2022empirical}, long-form QA \citep{krishna2021hurdles}, and document-level retrieval \citep{callan1994passage}. These challenges stem from their limited effective context window size during the pretraining process, posing new challenges in generalizing over long contexts.

\vspace{-1mm}
A straightforward approach is to continually pre-train or fine-tune these models on long context input \citep{fu2024data}. However, expanding the context window usually results in a quadratic increase in computational and memory costs. According to the training setup in \citep{fu2024data}, extending the Llama-2 7B model's context window from 4k to 80k using 8 A100 GPUs (80G each) takes five days. The costs of resources and time increase significantly for larger models and more extended training periods. 
In addition to the methods mentioned, there are techniques aimed at extending the length of the context window more efficiently during fine-tuning, including PI \citep{chen2023extending}, Yarn \citep{peng2024yarn}, and LongLoRA \citep{chen2024longlora}. However, these techniques still require full-length fine-tuning, meaning they must fine-tune with the context of the target length, which is both memory- and time-intensive. Meanwhile, Randomized Positional Encoding Scheme \citep{ruoss-etal-2023-randomized} and PoSE \citep{zhu2023pose} simulate longer inputs within a fixed window by adjusting position indices, allowing LLMs that are trained on shorter contexts to be extended to longer context windows.
However, randomized position embeddings in \citep{ruoss-etal-2023-randomized}  disrupt local sentence structures by exaggerating the dependency lengths between neighboring tokens. PoSE, on the other hand, only considers two chunks to mimic the position index, consistently omitting longer dependencies in the sequence. This distortion creates a significant generalization gap in understanding token relationships across the sequence when extending LLMs to a long context window.

\begin{figure*}
    \centering
    \includegraphics[width=0.9\linewidth]{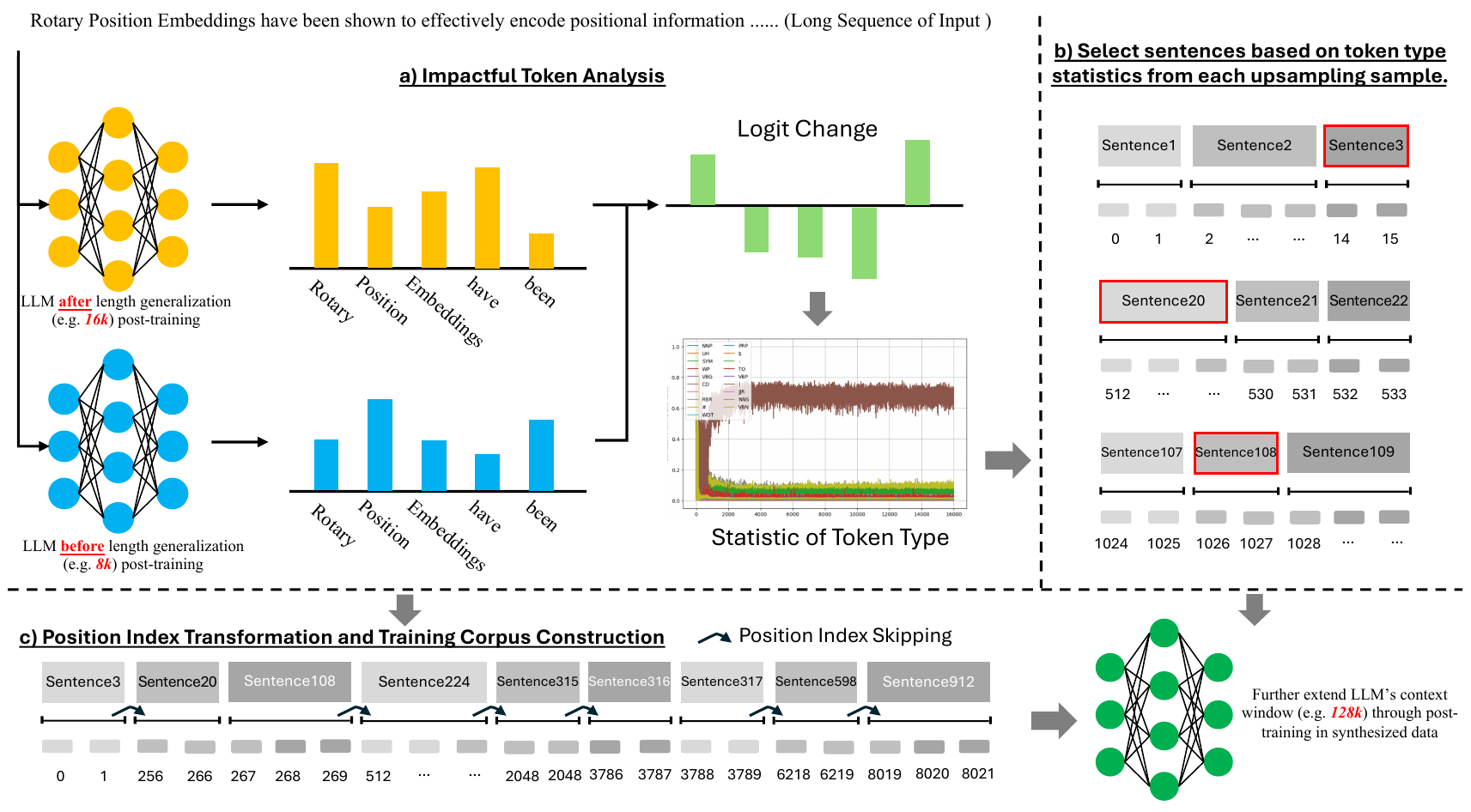}
    \vspace{-4mm}
    \caption{Method Overview}
    \label{overview}
    \vspace{-6mm}
\end{figure*}

To address the aforementioned issues and further uncover the potential of efficient training for long-context generalization in LLMs, 
we present \textbf{LongRecipe}, an efficient framework designed to enhance long-context capabilities in models. 
Long context generalization depends on token distances set by position indexs, which are then combined with token representations. LongRecipe  is primarily focused on optimizing the learning process by efficiently handling both position indexs and token representations.
Our framework introduces Impactful Token Analysis to identify tokens that significantly influence long-text training. By focusing on these tokens, we extract shorter segments from long-text corpora, reducing text length while preserving key information. We then apply Position Index Transformation to simulate long-sequence positional indices using these shortened texts, extending the model's ability to handle long sequences without needing actual long texts. Additionally, we implement training optimizations — pretraining data replay and model merging — to enhance the model's long-text processing capabilities.
As illustrated in Figure~\ref{overview}, LongRecipe compares the logits of output tokens from an untuned LLM with those from a tuned LLM within a longer context. This reveals significant token logit changes from long context generalization training. Sentences or paragraphs with these tokens are selected, upsampled, and segmented with continuous positional indices, then used to train the LLM, effectively extending its context window.
This method efficiently captures key changes in long-context training while improving training efficiency by streamlining samples. The position index transformation also sharpens the model's understanding of long-range dependencies and sequences in extended texts.

To validate the effectiveness of LongRecipe, we conduct the empirical evaluation with Llama3-8B, Mistral-7B, Qwen2-7B on Multi-Needle In A Haystack~\citep{gkamradt2023needle}, RULER \citep{hsieh2024ruler}, LongBench \citep{bai2023longbench} and Loogle \citep{li2023loogle}.
Applied with LongRecipe, we can extend the effective context window of an open-source LLM from 8k or 32k to 80k or 128k. The experimental results demonstrate that LongRecipe achieves an average improvement of approximately 5.5\% across four metrics in three types of models, with context windows 80k and 128k.
Additionally, using as little as 30\% of the tokens with around 1/8 of the GPU computational resources can achieve nearly the same performance as full context window training. Currently, we can extend an open-source LLM's context window from 8k to 128k, matching GPT-4-Turbo's performance with just one day of training on a single H100 GPU.
Furthermore, we test the performance of our method in general tasks, including MMLU \citep{hendrycks2021measuring}, GSM8K \citep{DBLP:journals/corr/abs-2110-14168}, and HumanEval \citep{chen2021evaluating} to assess if our method impacts LLMs' general abilities, showing it largely preserves their original performance. To summarize, our contributions are as follows:
\vspace{-2mm}
\begin{itemize}
    \item We introduce \textbf{LongRecipe}, leveraging impactful token analysis and position index transformation to fully uncover the potential of efficient training for long context generalization.
    
    \item LongRecipe uses training strategies of
    the pretraining data replay and model merging to enable LLMs to preserve the original foundational abilities and enhance long context generalization ability stably.
    
    \item  Experiments conducted on context lengths from 8k or 32k to 80k or 128k of three types of LLMs validate the effectiveness of LongRecipe.
\end{itemize}

\section{Preliminary}

 The approach that is widely used in previous pre-trained language models such as BERT~\citep{devlin2018bert} is to add position embedding vectors to word embedding vectors directly. For a sequence of tokens represented as $w_1, w_2, \cdots, w_L$, with their corresponding embeddings $\mathbf{x}_1, \mathbf{x}_2, \cdots, \mathbf{x}_L$, let $\mathbf{p}_1,\mathbf{p}_2,...\mathbf{p}_L$ be absolute position embedding, the position encoding of query($\mathbf{q}$) and key($\mathbf{k}$) are $\mathbf{q}_m = W_q(\mathbf{x}_m + \mathbf{p}_m) $ and $\mathbf{k}_n = W_k(\mathbf{x}_n + \mathbf{p}_n)$.  Then the unnormalized attention scores are calculated by dot-producting two vectors: $score(\mathbf{q}_m,\mathbf{k}_n) = \mathbf{q}_m^T\cdot \mathbf{k}_n$. 

Rotary Position Embedding (RoPE)~\citep{SU2024127063} is proposed to integrate relative positional information by modulating the query and key vectors in the attention mechanism. Let $D$ denote the dimension of hidden layers, the transformations applied are as follows:
\begin{equation}
    \mathbf{q}_m = W_q\mathbf{x}_m \cdot e^{i m \theta}, \quad \mathbf{k}_n = W_k\mathbf{x}_n \cdot e^{i n \theta},
\end{equation}

where $W_q$ and $W_k$ are $|D|\times|D|$ projection matrices, $m$ and $n$ are the positions of the tokens, and $\theta$ is a constant that adjusts the rotation based on token positions. 
$$\theta_i = 10000^{\frac{-2i}{D}}$$
RoPE operation on $\Vec{q} = W_q\mathbf{x}_m$ results $\mathbf{q}_m = $:
\begin{equation*}
\footnotesize{
\begin{bmatrix}
q_0 \\
q_1 \\
\vdots \\
q_{D-2} \\
q_{D-1}
\end{bmatrix}
 \otimes \begin{bmatrix}
\cos m\theta_0 \\
\cos m\theta_0 \\
\vdots \\
\cos m\theta_{\frac{D}{2}-1} \\
\cos m\theta_{\frac{D}{2}-1}
\end{bmatrix}  +  \begin{bmatrix}
q_1 \\
q_0 \\
\vdots \\
q_{D-1} \\
q_{D-2}
\end{bmatrix} \otimes \begin{bmatrix}
-\sin m\theta_0 \\
\sin m\theta_0 \\
\vdots \\
-\sin m\theta_{\frac{D}{2}-1} \\
\sin m\theta_{\frac{D}{2}-1}
\end{bmatrix}
}
\end{equation*}
The real part of the inner product between $\mathbf{q}_m$ and $\mathbf{k}_n$ captures the relative positional information, facilitating the model's understanding of token distances.
\section{Related Works}

\textbf{Position Encoding} Various position encoding methods have been proposed to perform extrapolation such as ALiBi~\citep{press2022train}, xPos~\citep{sun-etal-2023-length}, KERPLE~\citep{chi2022kerple}. 
RoPE~\citep{SU2024127063} and CoPE \citep{golovneva2024contextual}, the most widely used one, introduces a more complex mechanism.

\textbf{Efficient Pretraining or Fine-tuning Methods} Position Interpolation (PI)\citep{chen2023extending} downsizes position indices of long text to the original window size. NTK Interpolation\citep{peng2023ntkaware} adjusts rotation speed for small positions and linear interpolation for large ones. YaRN~\citep{peng2024yarn} improves NTK Interpolation with NTK-by-parts scaling to accommodate different RoPE features. Resonance RoPE~\citep{wang2024resonance} refines RoPE features with integer wavelengths, improving upon YaRN for better out-of-distribution position recognition. LM-Infinite~\citep{han2024lminfinite} encodes absolute positions for starter tokens and masks middle tokens, retaining relative positions for rare tokens. Randomized Positional Embedding~\citep{ruoss-etal-2023-randomized} simulates long text input with shorter texts by randomly selecting position indices. PoSE~\citep{zhu2023pose} uses a fixed context window, dividing it into chunks with skipping bias terms, enabling adaptation to all positions within the target length. LongLoRA~\citep{chen2024longlora} replaces ordinary attention with shift short attention. Temp Lora~\citep{wang2024greater} integrates context details into a temporary Lora module, incrementally trained with previously generated text. SelfExtend \citep{jin2024llm} and DCA \citep{an2024training} convert the attention computation for long sequences into chunk-based modules to achieve the training-free extension.

\section{Methodology}
\label{method}

\subsection{Impactful Token Analysis}

As shown in Figure~\ref{overview} (a) and (b), consider a base large language model $H$ with a context window size $ L $. This model is further trained to extend its context window to $ L' $, resulting in a new model denoted as $ H' $. Using the LongRecipe methodology, we can calculate the logit offset for each token by comparing the differences between the logits produced by $ H $ and $ H' $. We then identify the token types with the most significant changes in logits to serve as anchors for selecting sentences containing these token types, which are then used for upsampling.

Formally, for each token $ t $, we condition both the base model $ H $ and the extended model $ H' $ on the preceding prompt $ x<t $ to obtain the logit probability scores $ S_H(t \mid x<t) $ and $ S_{H'}(t \mid x<t) $, respectively. These scores represent the final unnormalized logits from the language modeling head over the vocabulary. The distribution of logit probability changes is then given by:
\begin{align}
     p(X_t \mid x<t) 
    = \text{softmax}\left[S_{H'}(X_t \mid x<t) - S_H(X_t \mid x<t)\right]
\end{align}

To formally describe the process of selecting token types with the most significant logit changes, we define a significance score $ \Delta(t) $ for each token type $ t $ as:
\begin{equation}
    \Delta(t) = \sum_{i=1}^{N} \left| S_{H'}(X_t^{(i)} \mid x<t^{(i)}) - S_H(X_t^{(i)} \mid x<t^{(i)}) \right|
\end{equation}

where $ N $ represents the total number of samples. We then rank all token types by their significance score $ \Delta(t) $, and select the token types with the highest scores as anchors. The selected tokens are used to identify and upsample sentences that contain these tokens.

Intuitively, we aggregate the distributions across all samples to derive the statistics of token types whose logit probability changes are most significant. We select the top 20\% of tokens based on their significance scores $\Delta(t)$ at each position (e.g., the $i$-th token in the sample). We then calculate the frequency of each token type (part-of-speech).

Subsequently, for a given sample, we first remove sentences that do not contain these token types, which generally constitute a significant portion of the total sentences. Then, from the remaining sentences, we select a fixed number of tokens to use for further training.

\subsection{Position Index Transformation}
Refer to Figure~\ref{overview} (c), we aim to utilize the current data with context window $L$ to enable the model with larger input context length $\hat{L}$ by further continual pretraining in the data with synthesized position indices. Let $S$ be the original sequence. We define a function $\textbf{seg}: S \rightarrow \{s_1, s_2, \ldots, s_N\} $ that partitions $S$ into $N$ segments, where each segment $s_i$ can be either a sentence or a paragraph, for $ 1 \leq i \leq N $. The function $\textbf{seg}$ satisfies the following conditions:
\begin{equation}
    S = s_1 \cup s_2 \cup \cdots \cup s_N  
\end{equation}
The union of all segments reconstructs the original sequence and segments are disjoint:
\begin{equation}
    s_i \cap s_j = \emptyset \quad \text{for all} \quad i \neq j  
\end{equation}
To vary the spacing between each segment, we will randomly skip some position indices from 0 to $M$, where $M$ is a parameter of our method. When $M=0$, the position indices of the two segments will be continuous.

We start by defining $\textbf{pos}(s_i)$ as the position index of the first token of segment $s_i$. For each segment, the position indices are sequentially increased by 1 for each token within that segment. The position index of the first token in the first segment is set to 0, i.e., $\textbf{pos}(s_1) = 0$.

For subsequent segments, we introduce a random skip represented by a function $g(s_i)$ which takes values from ${0, 1, \ldots, M}$. This function represents the gap before the start of segment $s_i$ and is determined randomly for each segment. Thus, the position index of the first token of segment $s_i$, for $i \geq 2$, can be defined recursively as follows:
\begin{equation}
    \text{pos}(s_i) = \text{pos}(s_{i-1}) + |s_{i-1}| + g(s_i) + 1
\end{equation}
Where $|s_{i-1}|$ represents the number of tokens in segment $s_{i-1}$. We repeat this process until the position index of the last token of the last segment $s_N$ does not exceed $\hat{L}$.

To achieve comprehensive coverage of the target context window, we re-sample both the length and skipping term of every chunk for each training example.

\subsection{Training Optimization Strategies}

When we extend the effective context window of LLMs, we also want to enable the LLMs with strong general abilities within their original context window. Therefore, we explore the below two training optimization strategies to achieve that. 

\textbf{Pretraining Data Replaying} \quad
In this module, we address the issue of maintaining a model's general capabilities during post-training by employing a Pretraining Data Replay strategy. Specifically, we define two datasets: $\mathcal{D}_1$, which represents the original pretraining data, and $\mathcal{D}_2$, which is a replay dataset derived from $\mathcal{D}_1$. Both $\mathcal{D}_1$ and $\mathcal{D}_2$ share the same distribution.

The replay dataset $\mathcal{D}_2$ is used for further training after the model undergoes long-sequence extension training. This process is intended to help the model recover and reinforce its general capabilities that may have been affected during the length extension training. Formally, during the replay phase, the model is trained on $\mathcal{D}_2$ to restore and enhance its generalization abilities: $\Theta_{\text{replay}} = \text{Train}(\Theta_{\text{extended}}, \mathcal{D}_2)$. Here, $\Theta_{\text{extended}}$ represents the model after it has undergone long-sequence extension training, and $\Theta_{\text{replay}}$ is the model after the replay phase using $\mathcal{D}_2$. 

\textbf{Model Merging} \quad
To maintain the general abilities of original LLMs trained in short context window, we utilize a model merging technique to integrate the capabilities of two distinct models: one that is the original model without context window extension ($\Theta^{(o)}$) and another that is trained with longer context and pretraining data replaying ($\Theta^{(replay)}$). 
We use two hyperparameters $\lambda_1$ and $\lambda_2$ to retains the general abilities of original models and the long context generalization of tuned model. The merged model can be represented by the following equation:
\begin{equation}
    \Theta_{\text{merge}} = \lambda_1 \Theta^{(o)} + \lambda_2 \Theta^{(replay)}
\end{equation}

\vspace{-3mm}
\section{Experimental Setup}

\subsection{Baselines}

\textbf{Full-length Text Training (FLT)}. We train the LLMs using a corpus that contains the full target context length. This approach serves as a baseline for comparing the performance and observing any potential loss when applying our method.

\textbf{Randomized Positional Encoding Scheme (RPES)} \citep{ruoss-etal-2023-randomized} simulates the positions of longer sequences and randomly selects an ordered subset to match longer length.

\textbf{Positional Skip-wisE (PoSE)} \citep{zhu2023pose} simulates long inputs using a fixed context window. It divides the original context window into two chunks and applies distinct skipping bias terms to manipulate the position indices of each chunk. These bias terms and the lengths of the chunks are changed for each training example, enabling model to adapt to all positions within the target length.


\vspace{-2mm}
\subsection{Dataset and Evaluation}

\textbf{Dataset for Training} \quad We use the dataset in the work \citep{fu2024data} as training set. 
The dataset derived from SlimPajama \citep{SlimPajama}, incorporates domain balancing and length upsampling.
This dataset includes 80k samples and 128k tokens for each, we use 10k samples in the experiments for all baselines.

\textbf{Benchmarks of Long Context Generalization} \quad
The \textbf{Needle In A Haystack (NIAH)} framework~\citep{gkamradt2023needle} tests LLMs' ability to retrieve hidden information by embedding a "needle" (fact) within a "haystack" (long document). As the current LLMs can almost perform perfectly in single-needle retrieval task, we use more challenging multi-needle retrieval task to evaluate LLMs' ability, namely NIAH(M).
\textbf{RULER}~\citep{hsieh2024ruler} offers flexible sequence lengths and task complexities with 13 sub-task categories, including retrieval and question answering. \textbf{LongBench}~\citep{bai2023longbench} is the first bilingual benchmark for long context understanding, featuring 21 tasks in six categories.
We supplement more details about these benchmark in Appendix~\ref{Dataset}.

\textbf{Datasets for Assessment of Fundamental Abilities of LLMs} \quad
We use three benchmarks to test if the continual pretraining process affects LLMs' fundamental abilities within their original context length.
\textbf{MMLU} covers 57 subjects across STEM, the humanities, the social sciences, philosophy, law,  medicine and more \citep{hendrycks2021measuring}.
\textbf{GSM8K} \citep{DBLP:journals/corr/abs-2110-14168} is a benchmark of math problems. 
\textbf{HumanEval} \citep{chen2021evaluating} is a code problem solving dataset.

\subsection{Setup}
\vspace{-1mm}
\textbf{Long Context Training} \quad
We train the LLMs using samples with 30\% of the extended context window length and optimize efficiency with FlashAttention 2 \citep{dao2022flashattention} and DeepSpeed Zero 3 \citep{aminabadi2022deepspeed}.
To further train the LLMs with longer context window, we utlize Accelerator of Huggingface \citep{accelerate} and  Sequence Parallel technique (e.g. DeepSpeed-Ulysses \citep{jacobs2023deepspeed} and Ring Attention \citep{liu2023ring, ZigZag}) to optimize the GPU memory demands. 
More details including RoPE scaling, Batch Size, Hours to Train and others are in Appendix~\ref{Setups}.

\textbf{Pretraining Data Replay} \quad We use WizardLM-evol-instruct-70k \citep{luo2023wizardmath} and Magicoder-OSS-Instruct-75K\citep{wei2024magicoder}, totally with 68M tokens. Based on the findings in \citep{yang2024model}, replaying 5\% to 10\% of the post-training dataset is considered the optimal configuration. For our setup, we use a batch size of 96, a learning rate of 5e-6, and a decay rate of 0.1.
 \textbf{Model Merging} \quad We set $\lambda_1$ and $\lambda_2$ as 0.5, hence it would be the average weight for model merging.

\textbf{LLMs} \quad We test various LLMs to evaluate our approach, including Llama3 \citep{llama3-1}, Mistral \citep{misral7bv3}, Qwen2 \citep{qwen2}, GPT-4 \citep{openai2023gpt4}, Gemini-1.5-Pro \citep{gemini} and others. Information about all models is in Appendix~\ref{Model}.

\vspace{-2mm}

\section{Experimental Performance}

\vspace{-3mm}
\begin{table*}[htb]
\footnotesize
\centering
\resizebox{\textwidth}{!}{
\begin{tabular}{l|ll|ccc|ccc} 
\toprule
    \multirow{2}{*}{\bf Model} &\multirow{2}{*}{\bf Length} &\multirow{2}{*}{\bf Method} & \multicolumn{3}{|c}{\bf Long Context Generalization} & \multicolumn{3}{|c}{\bf General Abilities} \\
    
    & & &  NIAH(M) & RULER &LongBench &  MMLU & GSM8K  &  HumanEval   \\ 

\midrule
    \multirow{7}{*}{Llama3-8B-I} 

    &\multirow{1}{*}{8k} 
    & Base Model &- &- &- &\textbf{65.7} &\textbf{71.4}  &\textbf{37.5} \\
    \cdashline{2-9}
    \noalign{\vskip 0.5mm}
    &\multirow{4}{*}{80k} 
    & \cellcolor{gray!20}FLT &\cellcolor{gray!20}82.3 &\cellcolor{gray!20}75.7 &\cellcolor{gray!20}28.1 &\cellcolor{gray!20}62.2 &\cellcolor{gray!20}54.5 &\cellcolor{gray!20}32.7 \\
    & &RPES &71.8 &71.4 &\textbf{27.9} &61.4 &53.1 &15.4 \\
    & &PoSE &68.8 &69.9 &27.2 &62.6 &58.2 &25.6  \\
    & &LongRecipe &\textbf{78.9} &\textbf{74.5} &26.9 &63.0 &57.9 &29.3 \\

    \cdashline{2-9}
    \noalign{\vskip 0.5mm}

    &\multirow{4}{*}{128k} 
    & \cellcolor{gray!20}FLT  &\cellcolor{gray!20}73.2 &\cellcolor{gray!20}75.8 &\cellcolor{gray!20}26.4 &\cellcolor{gray!20}58.3 &\cellcolor{gray!20}50.9 &\cellcolor{gray!20}16.5 \\
    && RPES  &72.7 &71.5 &\textbf{27.3} &59.2 &46.0 &16.8 \\
    & & PoSE &80.1 &75.3 &26.7 &61.9 &51.1 &21.1 \\
    & &LongRecipe  &\textbf{82.6} &\textbf{76.0} &25.5 &62.1 &54.9 &24.2 \\

\midrule
    \multirow{7}{*}{Mistral-7B} 
    &\multirow{1}{*}{32k} 
    & Base Model &- &- &- &\textbf{55.7} &\textbf{28.4} &\textbf{31.1} \\
    \cdashline{2-9}
    \noalign{\vskip 0.5mm} 
    &\multirow{4}{*}{80k} 
    & \cellcolor{gray!20}FLT &\cellcolor{gray!20}43.0 &\cellcolor{gray!20}57.4 &\cellcolor{gray!20}17.7 &\cellcolor{gray!20}52.6 &\cellcolor{gray!20}25.2 &\cellcolor{gray!20}25.6 \\
    & &RPES &60.4 &65.1 &\textbf{25.6} &51.8 &27.4 &24.7 \\
    & &PoSE &64.7 &65.0 &23.6 &54.9 &29.4 &27.6  \\
    & &LongRecipe &\textbf{64.7} &\textbf{67.2} &22.5 &53.7 &28.0 &27.6 \\

    \cdashline{2-9}
    \noalign{\vskip 0.5mm}

    &\multirow{3}{*}{128k} 
    & RPES  &41.9 &52.5 &21.8 &52.8 &26.5 &24.8 \\
    & & PoSE &35.9 &46.3 &22.3 &53.4 &25.9 &22.6 \\
    & &LongRecipe  &\textbf{53.4} &\textbf{58.2} &\textbf{23.7} &53.1 &26.0 &24.2\\

\midrule
    \multirow{7}{*}{Qwen2-7B} 
    &\multirow{1}{*}{32k} 
    & Base Model &- &- &- &66.1 &58.3  &20.3 \\
    \cdashline{2-9} 
    \noalign{\vskip 0.5mm} 
    &\multirow{4}{*}{80k} 
    & \cellcolor{gray!20}FLT* &\cellcolor{gray!20}64.7 &\cellcolor{gray!20}69.5 &\cellcolor{gray!20}28.2 &\cellcolor{gray!20}\textbf{68.4} &\cellcolor{gray!20}\textbf{63.1} &\cellcolor{gray!20}\textbf{27.4} \\
    & &RPES &73.7 &68.9 &27.8 &65.7 &55.1 &16.0 \\
    & &PoSE &70.0 &66.7 &\textbf{28.2} &66.6 &58.9 &17.7 \\
    & &LongRecipe &\textbf{79.5} &\textbf{70.8} &25.7 &65.7 &57.2 &19.1 \\

    \cdashline{2-9}
    \noalign{\vskip 0.5mm}

    &\multirow{4}{*}{128k} 
    & \cellcolor{gray!20}FLT* &\cellcolor{gray!20}52.7 &\cellcolor{gray!20}51.3 &\cellcolor{gray!20}28.2 &\cellcolor{gray!20}\textbf{68.4} &\cellcolor{gray!20}\textbf{63.1} &\cellcolor{gray!20}\textbf{27.4} \\
    
    && RPES  &64.6 &64.6 &\textbf{26.8} &65.5 &56.1 &14.8  \\
    & & PoSE &58.5 &60.1 &27.7 &67.1 &58.2 &20.9 \\
    & &LongRecipe  &\textbf{65.8} &\textbf{64.8} &26.2 &65.9 &58.7 &17.3 \\

\midrule
    \multirow{12}{*}{Other LLMs} 
    &\multirow{12}{*}{128k} 
    &Llama3.1-8B  & 72.0 &69.8& 24.5 & 62.0 & 41.8 & 38.4 \\
    & &Yi-9B-200k & 65.7 &62.3&30.3 &42.5 & 51.3&21.3\\
    & &Yi-34B-200k  & 84.9&77.3 & 29.1  &76.3 &67.2 &23.2 \\
    \cdashline{3-9}
    & &Qwen2-7B-Instruct & 38.8 & 52.5 & 31.9 & 69.5 & 55.6 & 43.3\\
    & &Gradient-Llama3-8B  &89.6 & 78.4 &27.8 & 59.4 & 49.9 & 13.4\\
    & &Llama3.1-8B-Instruct &89.0 &77.7 & 30.0  &73.0 &84.5 &72.6 \\
    & &GLM4-9B-Chat-1M &90.2 &79.9 & 29.4 &74.7 &84.0 & 70.1\\
    & &Llama3.1-70B-Instruct &68.3 &66.6 & 42.8   &86.0 &95.1 &80.5 \\
    && Qwen2-72B-Instruct &83.4 &53.7 & 31.0 &84.2 &89.5 &64.6 \\
    & &Gradient-Llama3-70B & 79.2 &72.1 & 31.8  & 72.5 & 73.4& 33.5  \\
    & &GPT-4 & 76.2&81.2 & 50.0  &80.5 &93.0 &73.2 \\

    & &Gemini-1.5-Pro & 82.0 &94.4 &48.5   &81.9 &91.7 &71.9 \\

\bottomrule
\end{tabular}
}
\caption{Performance of different methods in long context generalization tasks and general abilities benchmarks. FLT* in Qwen2-7B denotes the Qwen2-7B base model combined with YARN and DCA methods for targeting the context window, as detailed in their technical report. In `Other LLMs' part, the models  above dashed line are the base model and blow are instruction tuned models. All the experiment results of FLT, RPES and PoSE are implemented by us.}

\label{result: main}
\end{table*}

\subsection{Long Context Generalization}
The LongRecipe method shows an average improvement of 6.6\% over RPES and 7.8\% over PoSE in the NIAH(M) task. In the RULER evaluation, LongRecipe outperforms RPES by 2.9\% and PoSE by 4.7\%. Especially, In the NIAH task, Llama3-8B-I (80k) shows a 10.1\% improvement with LongRecipe over PoSE. In the RULER task, Mistral-7B (128k) improves by 11.9\%.

Compared to the performance of current closed-source and open-source LLMs with a 128k context window, LongRecipe not only surpasses base models like Yi-9B, Llama3.1-8B, and the instruction model Qwen2-7B-Instruct but also achieves performance comparable to Gradient-Llama3-8B, which uses four times the tokens and full-length training. Additionally, LongRecipe approaches the performance levels of GPT-4.

\subsection{Maintaining General Abilities}
Table~\ref{result: main} shows that LLMs can nearly maintain their general abilities with short inputs, as seen by the minor performance drop in MMLU. Despite some remaining gaps in mathematical (GSM8K) and programming (HumanEval) abilities, the model merging and pretraining data replay strategy successfully restored approximately 75\% and 65\% of the original capabilities.

\subsection{Ablation Study}

\paragraph{Benefits of Impactful Token Analysis and Position Index Transformation}  As shown in Table~\ref{Imapactful_token_analysis}, the performance will drop significantly in NIAH(M) and RULER metrics when we randomly select sentence from long sequence (LongRecipe (Random T)) rather than using analyzed token pattern. Additionally, the application of Position Index Transformation can bring average 3.3\% improvement from PoSE to LongRecipe (w/o T).

\begin{table}[htb]
    \centering

    \begin{tabular}{c|ccccc}
        \toprule
         Method &  NIAH(M) &RULER &LongBench \\
         \midrule
         PoSE & 68.8 &69.9 &27.2  \\
         LongRecipe (w/o T) & 71.9 & 71.7 & 29.2  \\
         LongRecipe (Random T) & 70.1 & 69.8 & 27.1  \\
         LongRecipe & 78.9 &74.5 &26.4\\
         \bottomrule
    \end{tabular}
    \vspace{3mm}
    \caption{Performance of different ablation setings, LongRecipe (w/o T) uses the short exiting samples as PoSE and apply position index transformation on it. LongRecipe (Random T) select the sentence randomly from long sequence of sample and construct a new short samples. All experiments are based on Llama3-8B-instruct and we use 30\% tokens of 80k target context window.}
    \vspace{-1mm}
    \label{Imapactful_token_analysis}
\end{table}

\paragraph{Effect of Pre-training Data Replay and Model Merging for Maintaining General Abilities} In Table~\ref{replay and model merge},
comparing the models before and after replaying shows noticeable improvement, particularly on the GSM8K dataset. By further merging with the original model, we can enhance the model’s general capabilities, as seen in the MMLU performance (63\% vs. 65.7\%). Although there are still some gaps in mathematical (GSM8K) and coding (HumanEval) capabilities, the model merging and pretraining data replaying successfully recovers approximately 75\% and 65\% of the original abilities.

\begin{table*}[htb]
    \centering
    \small
    \begin{tabular}{l|ccc|ccc|ccc}
    \toprule
    \multirow{2}{*}{\bf Method} & \multicolumn{3}{c|}{\bf Before Replaying} & \multicolumn{3}{c|}{\bf After Replaying}& \multicolumn{3}{c}{\bf After Model Merging}\\
    \cline{2-10}
    \noalign{\vskip 0.5mm}
         & MMLU & GSM8K &HE & MMLU & GSM8K &HE& MMLU & GSM8K &HE\\
        \midrule
        FLT  & 58.1 &  39.7 & 20.9 & 58.2 &  47.2 & 17.5& 62.2 & 54.5 & 32.3\\
        RPES  & 54.0 &  33.6 & 15.2 & 59.7 &  46.7 & 3.3& 61.8 &  53.9 & 12.6\\
        PoSE  & 58.1 &  39.1 & 17.0 & 60.7 &  49.5 & 5.7& 62.6 &  58.2 & 25.6\\
        LongRecipe  & 58.6 &  42.7 &20.1 & 62.1 &50.9 & 6.7 &  63.0 & 57.9& 29.3 \\
    \bottomrule
    \end{tabular}
    \vspace{2mm}
    \caption{Performance of different stages in long context generalization training, pretraining data replaying and model merging. HE represents HumanEval. All experiments are conducted using the Llama3-8B-instruct model, with 30\% of tokens utilized within an 80k token target context window.
}
    \label{replay and model merge}
\end{table*}

\textbf{Performance Comparison Based on Various Number of Tokens}

\begin{figure}[htb]
    \centering
    \begin{minipage}{0.38\textwidth}
        As the number of tokens per sample increases, the performance of each sample improves consistently. However, the benefit gained from increasing the number of tokens (i.e., extending the context length) diminishes. Even we increase the token ratio from 30\% to 100\%, only around 1\% improvement can be obtained.        
        This is particularly evident in the results of Llama3-8B for a 128-token context window, as shown in Table~\ref{result: main}, where we achieve even better performance than FLT with 100\% of the tokens.
    \end{minipage}%
    \hfill 
    \begin{minipage}{0.55\textwidth}
        \centering
        \begin{tabular}{c|ccccc}
            \toprule
             Ratio &  NIAH(M) &RULER &LongBench\\
             \midrule
             10\% & 65.3 & 67.4 & 29.6 \\
             20\% & 70.9 & 70.7 & 28.8 \\
             30\% & 78.9 & 74.5 & 26.9 \\
             40\% & 72.0 & 71.0 & 25.7 \\
             100\% &82.3 & 75.7 & 28.1 \\
             \bottomrule
        \end{tabular}
        \caption{We conduct context window extension experiments using Llama3-8B-I with an 80k token length. Starting from 10\%, which represents 8k tokens per sample, 20\% corresponds to 16k tokens, 30\% to 24k tokens, and 40\% to 32k tokens. The 100\% configuration utilizes entire long sample.}
        \label{Compression ratio}
    \end{minipage}
\end{figure}

\subsection{Analysis}

\paragraph{Distance Among Tokens and Continual Length of Segments} 
We suppose the effectiveness of position index transformation stems from improving distances among token while maintaining local information via continual segment. Therefore, we calculated the average distance among tokens and the average continuous segment length for different methods.

\begin{figure}[htb]
    \centering
    \begin{minipage}{0.50\textwidth}
        \centering
        \includegraphics[width=0.95\linewidth]{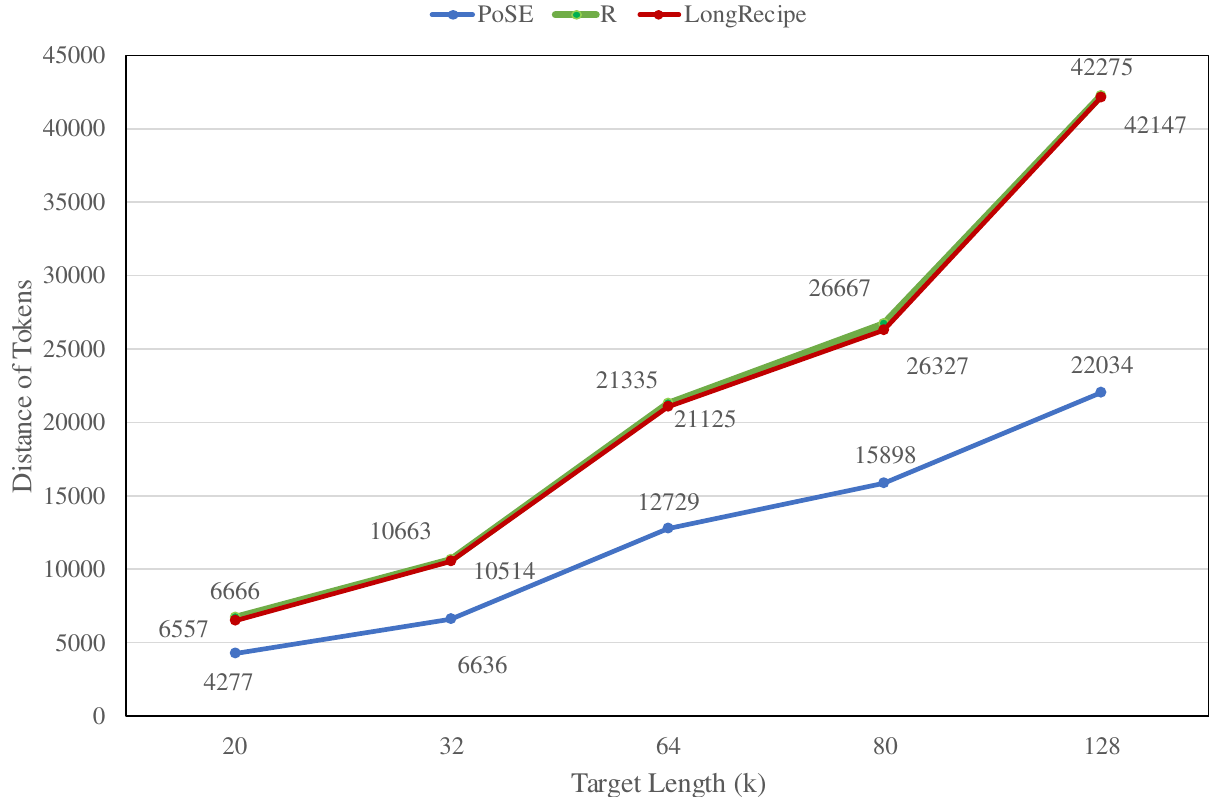}
        \vspace{-2mm}
        \caption{Comparison of average distance among tokens for different methods and context window.}
        \label{fig:distance}
    \end{minipage}%
    \hfill
    \begin{minipage}{0.45\textwidth}
         As shown in Figure~\ref{fig:distance}, the LongRecipe approach achieves approximately twice the token distance compared to PoSE in 128k setting. Additionally, LongRecipe maintains an average continuous segment length of 88, which helps the LLM recognize local dependency structures. In contrast, the average continuous segment length with RPES is nearly 0, greatly disrupting local sentence structures.
    \end{minipage}
    \vspace{-3mm}
\end{figure}

\begin{figure}[b]
    \centering
    \begin{minipage}{0.48\textwidth}
        
        This chart displays the frequency distribution and relative relationships of parts of speech for tokens with significant logits changes across different positions in the text. NUM (numerals) has the highest frequency, stabilizing around 0.4 throughout the text. In contrast, other parts of speech have significantly lower frequencies. For example, PRON (pronouns) and AUX (auxiliary verbs) have frequencies around 0.15, while CCONJ (conjunctions) and ADP (adpositions) have frequencies around 0.1. The frequency of NUM is approximately 2.67 times that of PRON and AUX and about 4 times that of CCONJ and ADP. These findings suggest that long-context tuning has varying effects on different token types, which further reinforces the motivation behind our method.
    \end{minipage}
    \hfill
    \begin{minipage}{0.47\textwidth}
        \centering
        \includegraphics[width=\linewidth]{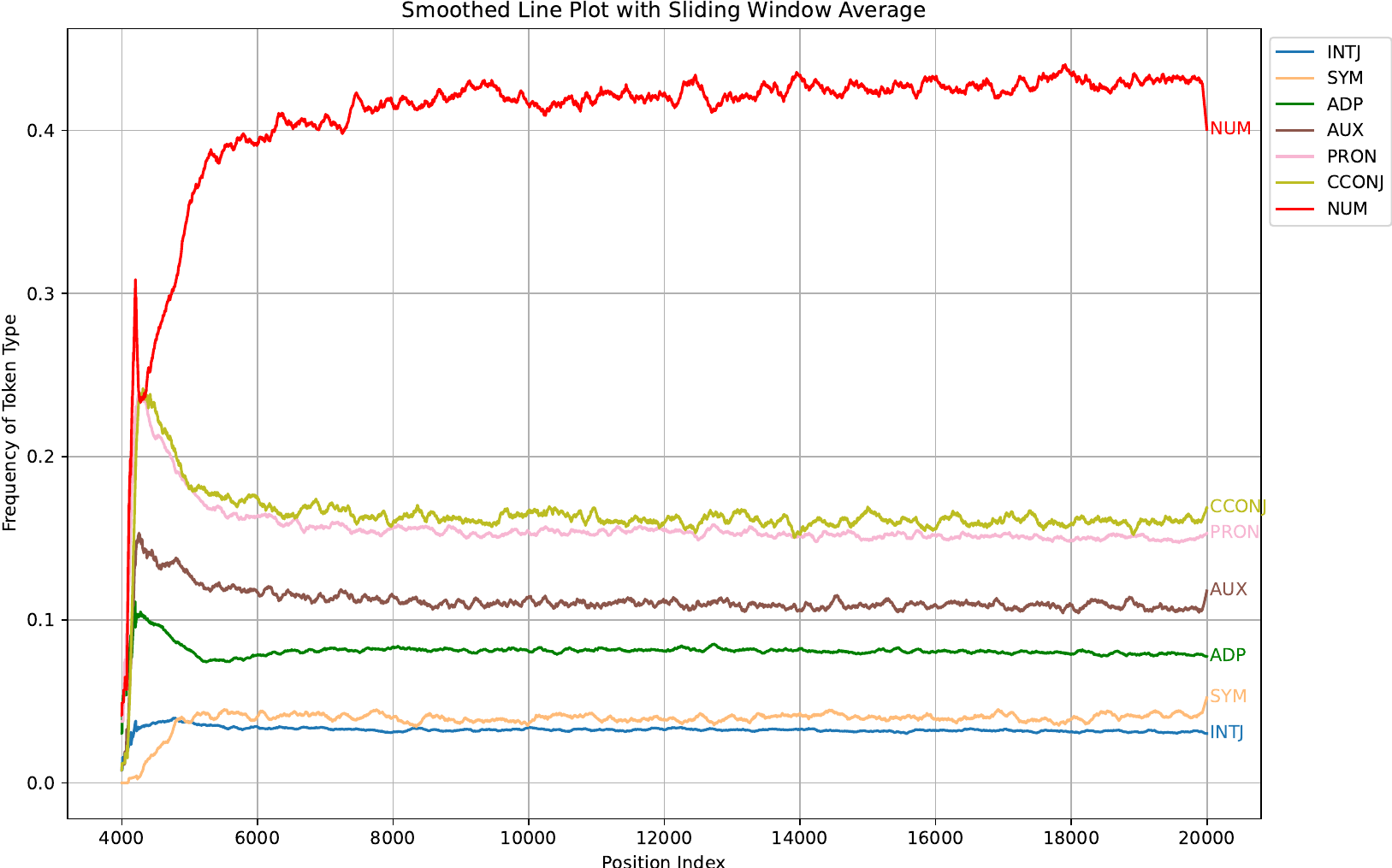}
        \caption{Frequency Distribution of Parts of Speech for Tokens with Significant Logits Changes Across Text Positions. INTJ (Interjection), SYM (Symbol), ADP (Adposition), AUX (Auxiliary), PRON (Pronoun), CCONJ (Conjunction), NUM (Numeral)}
        \label{token pattern}
    \end{minipage}
\end{figure}

\paragraph{Analysis of Token Type} In the LongRecipe approach, we first compare the change in token logits before and after tuning the long context window. We then select the top 20\% of tokens that exhibit the most significant change at each index. These selected tokens are further analyzed for their part of speech distribution patterns. The results of this analysis are presented in Figure~\ref{token pattern}.

\paragraph{Do Coherence and Cohesion Matter in Long Context Generalization?}

In this work, we select sentences from long samples based on analyzed token patterns, which may impact semantic coherence and cohesion. However, our current results based on LongRecipe can match or even surpass those from some full-length samples, suggesting that coherence and cohesion may not be as critical for long-context training. To further investigate this, we utilize the Long Dependency Score \citep{chen2024long} to assess the long dependencies in different datasets, which may be more crucial for long context training. After calculation, PoSE and RPES, which use the same existing short samples, achieved a score of 12.07, while the data constructed by LongRecipe and that concatenated from several short documents in FLT scored 17.88. Since the data used in FLT is concatenated rather than naturally occurring long context, the semantic quality is not satisfactory. LongRecipe does not significantly harm the long dependencies required for long-context training, even though it may influence coherence and cohesion to some extent.

Furthermore, during the pretraining process, the model primarily focuses on learning grammar and semantics. However, in the post-training phase on long texts, the model has already acquired grammar and semantic knowledge, so the focus may shift to capturing long dependencies among tokens. At this stage, it is possible to ignore certain tokens that are less important. These sentence might significantly influence overal semantic but contribute little to the learning of more complex attention patterns across longer sequences. Although this approach may slightly affect the model's general capabilities, the impact is minimal, and the model can quickly recover. More importantly, we can leverage LongRecipe method to achieve the efficient training for long context training.

\section{Conclusion}

In this work, we presented LongRecipe, a novel and efficient framework for extending the context window of  LLMs to enhance their performance on long-context tasks. By integrating impactful token analysis, position index transformation, and training strategies, LongRecipe effectively simulates long-sequence inputs while maintaining training efficiency. Our extensive experiments on various LLMs, with extended context windows in 80k to 128k, demonstrated that LongRecipe could achieve substantial improvements in long-context generalization with significantly reduced computational resources. Notably, the method requires only 30\% of the target context window size and cuts down training costs over 85\% compared to full-length post-training. Moreover, LongRecipe preserves the original capabilities of the LLMs in general tasks, ensuring a balanced enhancement of both long-range dependency understanding and foundational model performance.

\section{Limitation and Onging Work}

\paragraph{Supervised Fine-Tuning (SFT)} While our current post-training approach, based on instruction or base models, yields satisfactory performance in NIAH and RULER, the absence of SFT still creates a gap between our method and the state-of-the-art (SOTA) LLMs. Recently, the release of LongWriter \citep{longwriter} for long-context SFT presents a promising option for further enhancing our fine-tuning process.

\paragraph{Longer Context Generalization} The latest LLMs have pushed the boundaries of long-context capabilities to handle up to 1 million tokens, enabling users to input vast amounts of data. We plan to train and release models with 512k and 1M token capacities using the effective training strategies outlined in LongRecipe. This approach will further enhance the generalization of our method.

\clearpage
\bibliography{iclr2024_conference}
\bibliographystyle{iclr2024_conference}

\clearpage
\appendix

\section{Training Setup}
\label{Setups}
\vspace{-3mm}
\begin{table*}[htb]
\label{setup}
\fontsize{9pt}{10pt}\selectfont
  \begin{center}
    \begin{tabular}{l|cc|cc}
    \toprule
    
    \bf Model & \multicolumn{2}{c}{\bf Llama3-8B} & \multicolumn{2}{c}{\bf Qwen2-7B}   \\
    \midrule
    Extended Context Length & 80k & 128k & 80k & 128k \\
    Training Sample Length &24k &38.4k &24k &38.4k \\
    RoPE scaling (Dynamic NTK) &48.9M &131.5M &13.5M &13.5M \\
    RoPE factor (Dynamic NTK) &10 &16 &4 &4 \\
    Batch Size &96 &96 &96 &96 \\
    Steps &104 &104 &104 &104 \\
    Total Tokens &240M &384M &240M &384M \\
    Learning Rate &5e-5  &5e-5 &5e-5  &5e-5  \\
    \# GPUs and Type &1×A800/H100 &2×A800/H100 &1×A800/H100 &1×A800/H100 \\
    Total GPU Memory &56G &104G &64G &72G \\
    Total CPU Memory &148G &172G &168G &208G \\
    Hours to Train & 26/16 & 30/20 & 23/15 & 44/28\\
    \bottomrule
    \end{tabular}
  \end{center}
  \vspace{-1em}
  \caption{Training Configuration Details}
  
\end{table*}

\vspace{-3mm}
\section{Models}
\label{Model}

We select in total 15 models for evaluation and analysis. We assess two commercial close-source GPT-4 and Gemini-1.5, and 13 open-source models. 

\begin{table}[H]
\centering
\resizebox{0.99\linewidth}{!}{
\begin{tabular}{@{}lccl@{}}
\toprule
Model  & Size & Context Length & Huggingface~\citep{huggingface} / API \\
\midrule
GPT-4~\citep{openai2023gpt4}  & - & 128K & \texttt{gpt-4-1106-preview} \\
Gemini-1.5-Pro~\citep{gemini}  & - & 1M & \texttt{gemini-1.5-pro} \\
\midrule
Llama3-8B-I~\citep{llama3-1}  & 8B & 8K & meta-llama/Meta-Llama-3-8B-Instruct \\
Llama3.1-8B~\citep{llama3-1}  & 8B & 128K & meta-llama/Meta-Llama-3.1-8B \\
Llama3.1-8B-Instruct~\citep{llama3-1}  & 8B & 128K & meta-llama/Meta-Llama-3.1-8B-Instruct \\
Llama3.1-70B-Instruct~\citep{llama3-1}  & 70B & 128K & meta-llama/Meta-Llama-3.1-70B-Instruct \\
Qwen2-7B~\citep{qwen2}  & 7B & 128K & Qwen/Qwen2-72B-Instruct \\
Qwen2-7B-Instruct~\citep{qwen2}  & 7B & 128K & Qwen/Qwen2-7B-Instruct \\
Qwen2-7B-Instruct~\citep{qwen2}  & 72B & 128K & Qwen/Qwen2-72B-Instruct \\
Yi-9B-200k~\citep{yi}  & 9B & 200K & 01-ai/Yi-34B-200K \\
Yi-34B-200k~\citep{yi}  & 34B & 200K & 01-ai/Yi-34B-200K \\
Mistral-7B~\citep{misral7bv3}  & 7B & 32K & mistralai/Mistral-7B-Instruct-v0.3 \\
GLM4-9B-Chat-1M~\citep{glm4}  & 9B & 1M & THUDM/glm-4-9b-chat-1m \\
Gradient-Llama3-8B~\citep{gradient-llama3}   & 8B & 1M & gradientai/Llama-3-8B-Instruct-Gradient-1048k \\
Gradient-Llama3-70B~\citep{gradient-llama3}   & 70B & 1M & gradientai/Llama-3-70B-Instruct-Gradient-1048k \\

\bottomrule
\end{tabular}}
\vspace{2mm}
\caption{Information of evaluated and analyzed models in.}
\end{table}

\section{Pseudo Code for Position Index Transformation}

\begin{algorithm}
\caption{Position Index Transformation}
\begin{algorithmic}[1]
\State \textbf{Initialize:}
\State Initialize source length $L_s$ and target length $L_t$
\State Load dataset $\mathcal{D}$ with each sample having length $L_s$
\State \textbf{Position Index Transformation:}
\For{each sample $\textit{S}$ in $\mathcal{D}$}
    \State Split $\textit{S}$ into $\textit{N}$ sentences based on delimiters '. ! ? \textbackslash n'
    \State Initialize a list $\mathcal{L}$ of length $L_t$, filled with zeros
    \State Randomly select $\textit{N}-1$ distinct positions in $\mathcal{L}$
    \State Insert the first sentence at position 0 and each of the remaining sentences at the selected $\textit{N}-1$ positions in $\mathcal{L}$
    \State Flatten $\mathcal{L}$ by removing zeros, and the indices of the non-zero elements represent the new position indexes
\EndFor
\State \textbf{Save New Position Indexes}
\end{algorithmic}
\end{algorithm}

\section{Details about Long Context Benchmarks}
\label{Dataset}
\textbf{NIAH (M) and RULER}: For NIAH (M), we report the average score across three tasks in RULER: \textit{niah\_multikey}, \textit{niah\_multivalue}, and \textit{niah\_multiquery}. For RULER, we present the average score for all 13 subsets with Llama3-8B-Instruct and Mistral-7B-v0.3, and the average score for 12 subsets (excluding \textit{Variable Tracking}) with Qwen2-7B.

\textbf{LongBench}: For LongBench, we report the average score across all 21 subsets for the models.


\end{document}